\documentclass[runningheads]{llncs}

\usepackage{eccv}

\usepackage{eccvabbrv}

\usepackage[utf8]{inputenc}

\usepackage{graphicx}
\usepackage{booktabs}

\usepackage[accsupp]{axessibility}  

\usepackage{amssymb}
\usepackage{todonotes}
\usepackage{hhline}
\usepackage{verbatim}

\usepackage{hyperref}

\usepackage{orcidlink}

\usepackage{amsmath} 
\usepackage{booktabs}
\newcommand\headercell[1]{
   \smash[b]{\begin{tabular}[t]{@{}c@{}} #1 \end{tabular}}}

\usepackage{bbm}

\usepackage{makecell}

\begin{document}

\title{Tracking one-in-a-million: Large-scale benchmark for microbial single-cell tracking with experiment-aware robustness metrics}

\titlerunning{TOIAM}

\author{Johannes Seiffarth\inst{1,2}\orcidlink{0000-1111-2222-3333} \and
Luisa Blöbaum\inst{3}\orcidlink{0000-0002-3294-1695} \and Richard D. Paul\inst{4}\orcidlink{0000-0002-7433-0075}  \and Nils Friederich\inst{5,6}\orcidlink{0009-0003-9377-3140
} \and Angelo Jovin Yamachui Sitcheu\inst{5}\orcidlink{0009-0002-9384-6496} \and Ralf Mikut\inst{5}\orcidlink{0000-0001-9100-5496} \and Hanno Scharr\inst{4}\orcidlink{0000-0002-8555-6416}\and Alexander Grünberger\inst{3, 7}\orcidlink{0000-0002-7564-4957} \and Katharina Nöh\inst{1}\orcidlink{0000-0002-5407-2275}}

\authorrunning{J.~Seiffarth et al.}

\institute{Institute of Bio- and Geosciences, IBG-1: Biotechnology, Forschungszentrum Jülich GmbH, Jülich, Germany \and
Computational Systems Biology (AVT-CSB), RWTH Aachen University, Aachen, Germany,
\and Multiscale Bioengineering, 
Bielefeld University, Bielefeld, Germany
\and Institute for Advanced Simulation, IAS-8: Data Analytics and Machine Learning, Forschungszentrum Jülich GmbH, Jülich, Germany
\and
Institute for Automation and Applied Informatics, Karlsruhe Institute of Technology, Eggenstein-Leopoldshafen, Germany
\and Institute of Biological and Chemical Systems, Karlsruhe Institute of Technology, Eggenstein-Leopoldshafen, Germany
\and Microsystems in Bioprocess Engineering, Karlsruhe Institute of Technology, Karlsruhe, Germany
}

\maketitle

\begin{abstract}
  
  Tracking the development of living cells in live-cell time-lapses reveals crucial insights into single-cell behavior and presents tremendous potential for biomedical and biotechnological applications. In microbial live-cell imaging (MLCI), a few to thousands of cells have to be detected and tracked within dozens of growing cell colonies. The challenge of tracking cells is heavily influenced by the experiment parameters, namely the imaging interval and maximal cell number. For now, tracking benchmarks are not widely available in MLCI and the effect of these parameters on the tracking performance are not yet known.
  
  Therefore, we present the largest publicly available and annotated dataset for MLCI, containing more than $1.4$ million cell instances, $29$k cell tracks, and $14$k cell divisions. With this dataset at hand, we generalize existing tracking metrics to incorporate relevant imaging and experiment parameters into experiment-aware metrics. These metrics reveal that current cell tracking methods crucially depend on the choice of the experiment parameters, where their performance deteriorates at high imaging intervals and large cell colonies.
  Thus, our new benchmark quantifies the influence of experiment parameters on the tracking quality, and gives the opportunity to develop new data-driven methods that generalize across imaging and experiment parameters. The benchmark dataset is publicly available at \url{https://zenodo.org/doi/10.5281/zenodo.7260136}.
  
\end{abstract}

\section{Introduction}
\label{sec:intro}

\begin{figure}
    \centering
    \includegraphics[width=1\linewidth]{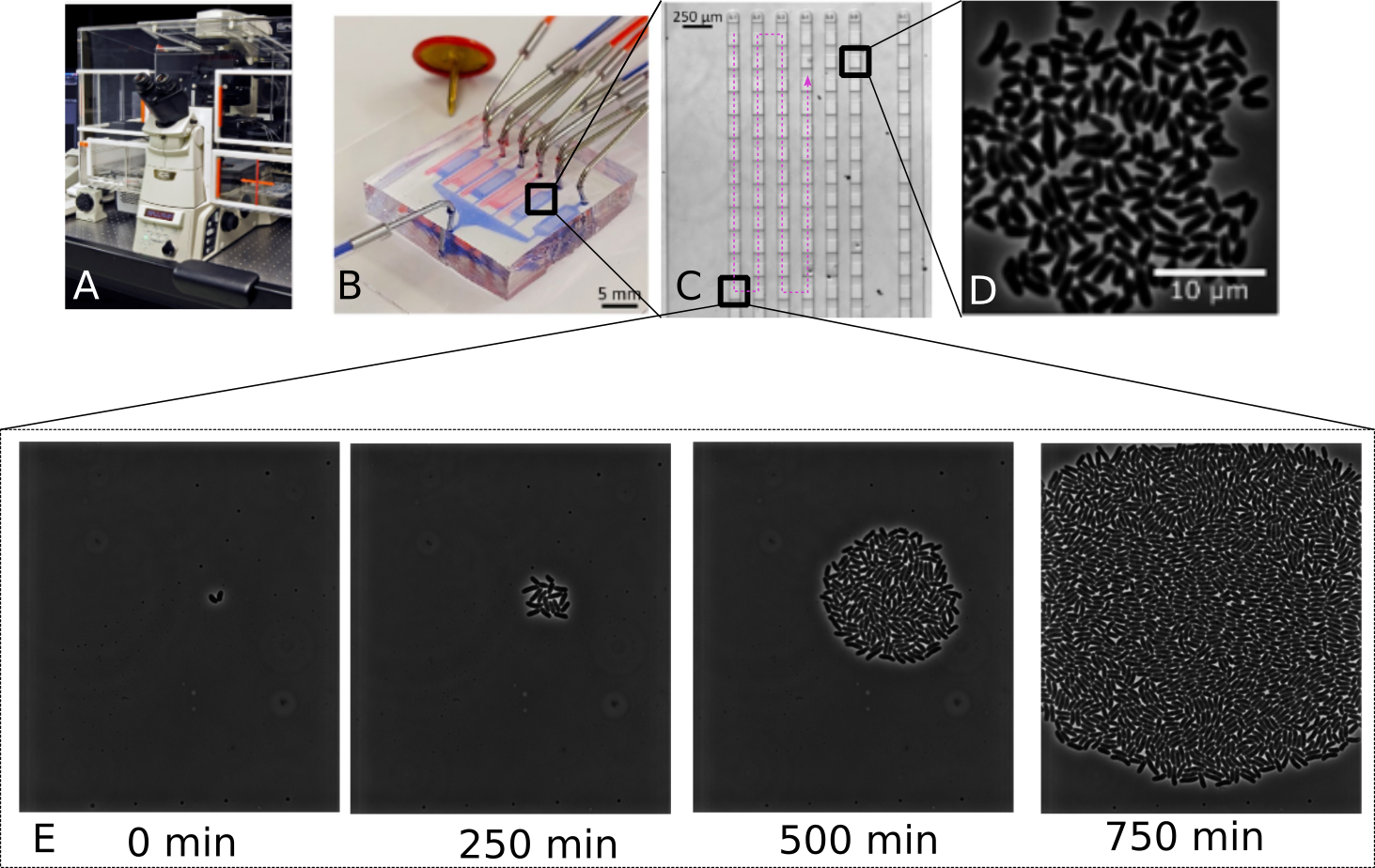}
    \caption{\textbf{Data acquisition in microbial live-cell imaging}. A microfluidic cultivation device is mounted to an automated microscope (A, B). The device contains hundreds to thousands of rectangular cultivation chambers (C) that are imaged one-by-one, moving the microscope stage and capturing images at a series of time-points (D). This imaging-and-movement loop, indicated by the purple line (C), is repeated with a pre-set imaging interval and leads to time-lapse recordings capturing the temporal development of the cell colony from a few cells up to several thousands (E). The images in (E) depict a $80 \times 90 \mu m$ region. The figure is adapted from Blöbaum \etal\cite{blobaum_quantifying_2024}. 
    }
    \label{fig:mlci}
\end{figure}

Detecting objects, segmenting their visual appearance into pixel-precise masks and tracking their movement through time is a fundamental challenge of computer vision providing crucial scene understanding necessary for autonomous driving \cite{geiger_vision_2013}, pedestrian management~\cite{pellegrini_youll_2009}, sports or robotics \cite{geiger_vision_2013}. Especially, in biomedical imaging, tracking the development of individual living cells allows gaining insights into the basic principles of life and diseases.
For instance, single-cell tracking allows studying virus infections~\cite{sivaraman_detecting_2011}, pathogenic bacteria~\cite{hockenberry_microbiota-derived_2021}, cell aging~\cite{oconnor_delta_2022}, and cell interactions~\cite{van_vliet_spatially_2018,helfrich_live_2015} at the single-cell level.
In particular, microbial live-cell imaging (MLCI) is a technology that performs high-throughput screening of the temporal developments of individual cells (see Figure \ref{fig:mlci}). Herein, living microbial cells are introduced into microfluidic chip devices and trapped within thousands of micrometer-sized microfluidic structures called cultivation chambers. Within these structures, the cells grow in monolayers while their temporal development is recorded using automated microscopy. The microscope scans the cultivation chambers one by one, takes an image and repeatedly performs this within a loop, recording a time-lapse that captures the temporal development of the independent cell colonies. As a result, a single MLCI experiment records dozens of time-lapses and produces hundreds of gigabytes of raw imaging data. Clearly, automated segmentation and tracking methods are essential to extract information from the time-lapse images and to gain insights into microbial colony development and single-cell behavior.

However, tracking living microbial cells presents unique challenges distinct from those usually encountered in general object tracking.
First, living microbial cells divide frequently, with division times ranging from a few minutes to hours. 
In MLCI experiments, few cells grow exponentially into dense and large colonies with up to thousands of cells captured in a single microscopy image (Figure~\ref{fig:mlci}E) while their total number is limited by the size of the cultivation chamber.
Second, microbial cells in phase-contrast microscopy are visually hard to distinguish, making it hard to track them by their appearance (see Figure~\ref{fig:mlci}D).
Third, the time-lapse recordings are affected by parameters such as the choice of the imaging interval between two consecutive phase contrast images and the number of concurrently monitored cultivation chambers. 
Moreover, both parameters are interdependent: lower imaging intervals usually simplify the tracking challenge but enforce shorter movement cycles of the microscope and, consequently, limit the number of concurrently monitored cultivation chambers.
Notably, the imaging procedure itself may influence the growth behavior of the cells. While phase contrast imaging is not considered to impact microbial growth, fluorescence imaging may lead to phototoxicity and -bleaching effects~\cite{hoebe_controlled_2007}.
Therefore, MLCI experiments are usually conducted with relatively low imaging rates. This leads to frequent cell divisions and larger cell displacement between consecutive frames. Consequently, we argue that MLCI not only needs robust and highly automated cell tracking, but also informed choices of experiment parameters, namely the imaging interval and the upper cell count limit.

In recent years, the rapid development in deep learning (DL) methods has driven segmentation and tracking methods~\cite{he_mask_2018,liu_cbnet_2019,qiao_detectors_2020,fang_eva_2022,zhang_fairmot_2021,zhang_bytetrack_2022,wang_smiletrack_2024,kirillov_segment_2023}. Moreover, the increased availability of large-scale datasets such as ImageNet~\cite{russakovsky_imagenet_2015}, KITTI~\cite{geiger_vision_2013}, CityScapes~\cite{cordts_cityscapes_2016}, SA-1B~\cite{kirillov_segment_2023}, and LAION-5B~\cite{schuhmann_laion-5b_2022} has shown to be a crucial driver for method development. Within the life sciences, we have seen a similar rapid development of methods. DL segmentation approaches have been developed~\cite{cutler_omnipose_2022,stringer_cellpose_2021,jeckel_advances_2020,upschulte_uncertainty-aware_2023,scherr_cell_2020}, driven by annotated datasets \cite{ulman_objective_2017,stringer_cellpose_2021,caicedo_nucleus_2019,edlund_livecelllarge-scale_2021,schwartz_caliban_2023,greenwald_whole-cell_2022}. For cell tracking, special methods have been developed that incorporate cell divisions\cite{schwartz_caliban_2023,jaqaman_robust_2008,theorell_when_2019,loffler_graph-based_2021,gallusser_trackastra_2024,loffler_embedtracksimultaneous_2022,maska_cell_2023}. However, public datasets for cell tracking \cite{ulman_objective_2017,schwartz_caliban_2023,anjum_ctmc_2020} primarily focus on microscopy images with eukaryotic cells. In contrast to microbial cells, eukaryotes usually show more distinctive visual features, less frequent cell division and the images contain fewer cell instances. For microbial cells, only few tracking datasets~\cite{oconnor_delta_2022} are available, making it difficult to train data-driven tracking methods and benchmarking suitable experiment parameters. Due to the lack of datasets, the importance of the experiment parameters for high-quality automated tracking has not been investigated before.

Therefore, we establish a novel benchmark for cell tracking in MLCI and extend existing tracking metrics with experiment parameters to quantify their effect on the tracking performance. Our contributions are threefold:
(1) we introduce a new annotated time-lapse dataset, recorded with low imaging interval for the segmentation and tracking of \textit{Corynebacterium~glutamicum} cells containing roughly $1.4$ million cell masks and about $14$k cell divisions. 
(2) We introduce experiment-aware metrics that extend existing metrics and incorporate the choice of the imaging interval and the maximum number of cells into the evaluation. 
(3) We evaluate state-of-the-art (SOTA) tracking methods using our devised metrics across a broad range of experiment parameters. 
We thereby show that the performance of SOTA tracking algorithms deteriorates, especially at lower imaging rates and higher cell counts. 
Notably, this fact that has not yet been quantified by the CTC community. 
Therefore, our benchmark represents a step forward to towards fully automated and robust data-driven microbial single-cell tracking and raises awareness about the importance of experiment parameters for cell tracking in MLCI experiments.

\section{Related Work}

\textbf{Benchmark datasets}. Benchmark dataset for cell segmentation cover different cell tissues, morphologies and imaging modalities~\cite{caicedo_nucleus_2019,edlund_livecelllarge-scale_2021,schwartz_caliban_2023,stringer_cellpose_2021}. The availability of additional tracking information for full time-lapse recordings is much less common. The cell tracking challenge combines datasets of various cell types and provides partial dense segmentation and full tracking information\cite{ulman_objective_2017,maska_cell_2023}. Schwartz \etal introduced the DynamicNuclearNet dataset containing roughly 600K segmented nuclei instances with roughly $2$k cell divisions\cite{schwartz_caliban_2023}. Van Vliet \etal provide a dataset of six genetic variants of \textit{Escherichia coli} in 39 time-lapse videos containing roughly $100$k cell instances and $9$k cell divisions. Anjum \etal introduce the CTMC challenge dataset contains roughly 2 million cell detections within $2.9$k cell tracks and $457$ cell division events~\cite{anjum_ctmc_2020}. Their segmentation annotation is restricted to bounding boxes and on average $13$ cells are visible within a microscope image.

\textbf{Tracking Methods}. In the predominant tracking-by-detection scheme, cells are first detected in the microscopy time-lapse and then linked across frames to build biologically valid cell tracks. Thus, tracking-by-detection is usually formulated as a graph problem, where nodes represent the cell detections at specific points in time and edges link cell detections through time. Therefore, Jaqaman \etal \cite{jaqaman_robust_2008} formulated tracking as a linear assignment problem (LAP) where cell detections are linked into segments, which are then linked to incorporate cell division events. Theorell \etal \cite{theorell_when_2019} extended this approach into multi hypothesis tracking (MHT), where cell linking costs are derived from biological models and tracking predictions are sampled using a particle filter approach. In contrast, Löffler \etal \cite{loffler_graph-based_2021} formulated a coupled minimum cost flow and correct segmentation errors during the tracking. All these methods require hand-tuned parameters to compute the costs of linking cells.

Data-driven approaches promise to derive these linking costs purely from training data. Ben-Haim \etal \cite{ben-haim_graph_2022} use a graph neural network (GNN) to predict cell linking costs and utilize features computed from contrastive visual embeddings. However, cell divisions are performed heuristically among the cells' neighborhood. Similarly, Schwartz \etal \cite{schwartz_caliban_2023} utilize a graph attention network to predict linking costs used for building a LAP. Gallusser \etal \cite{gallusser_trackastra_2024} extract cell detection features and use a transformer network to predict the linking costs. The tracking graph is constructed using a greedy scheme or optimizing an integer linear program (ILP) for minimizing linking costs. O'Connor \etal \cite{oconnor_delta_2022} predict the dense evolution of cell mask to the next time point for every cell instance using a U-Net architecture. Cells are then linked by their mask overlap. While Gallusser \etal and O'Connor \etal have been using microbial datasets, the other data-driven approaches have focused on tracking eukaryotic cells.

\textbf{Tracking metrics.} General object tracking metrics such as MOTA \cite{bernardin_evaluating_2008} and HOTA \cite{luiten_hota_2021} have been established for tracking a wide range of objects, but lack the consideration of object division crucial in cell tracking. Thus, specialized tracking metrics have been developed and established in cell tracking, especially by the cell tracking challenge (CTC) \cite{ulman_objective_2017}. Herein, we distinguish technical and biological tracking metrics. Technical tracking metrics such as the \textit{TRA} and \textit{LNK} are based on the Acyclic Oriented Graph Matching (AOGM)\cite{matula_cell_2015}. The AOGM is a weighted sum of costs for the minimal set of operations to transform the predicted segmentation and tracking into the ground truth segmentation and tracking. While the \textit{TRA} metric scores both segmentation and tracking errors, the \textit{LNK} metric solely rates errors in cell linking. Biological metrics are motivated by biological events that are of special interest. For instance, the complete tracks score (CT) measures the number of completely correctly reconstructed cell tracks from the first detection of a cell to its division or disappearance. Moreover, the mitotic branch correctness (MC) rates the quality of reconstructing cell division events.

\section{Benchmark Dataset}

In this work, we present a new MLCI benchmark dataset for microbial single-cell tracking, briefly termed 'Tracking one-in-a-million' (TOIAM). The dataset consists of microscopy time-lapses of growing \textit{C.~glutamicum} that show a characteristic 'snapping' division behavior, adding another challenge to the cell tracking. The images are recorded using phase contrast imaging at low imaging intervals of one image per minute. The recorded microscopy frames were annotated with cell segmentation masks and tracking information using a semi-automated workflow. We highlight the special characteristic of microbial datasets that are crucial to consider for robust cell tracking.

\subsection{Data acquisition}

MLCI experiments are usually carried out in three steps~\cite{tauber_how_2021}: First, the microbial organism is cultivated in a so-called preculture until reaching a certain biomass measured by the optical density (OD). Second, the cell suspension is introduced into a microfluidic cultivation chip, trapping individual cells within the cultivation chambers. Third, medium supply is connected to the microfluidic chip and the imaging routine is started recording hundreds of cultivation chambers at a specific imaging interval.

In our case, we cultivated \textit{C.~glutamicum} (ATCC 13032) in BHI-medium at $30~^\circ$C. From an overnight preculture, the main culture has been inoculated the next day starting with an OD600 of $0.05$ and grown at $120$~rpm to an OD600 of $0.25$. A microfluidic chip has been fabricated according to Täuber \etal \cite{tauber_dmscc_2020}, and fixed to the microscope’s stage. The main culture cells were transferred to monolayer cultivation chambers (height of $720$~nm) on the microfluidic chip within the inoculation procedure. Constant medium flow through the microfluidic device has been provided by pressure driven pumps with a pressure of $100$~mbar on the medium reservoir.

The time-lapse phase contrast images of five cultivation chambers have been recorded every minute using an inverted microscope (Nikon Eclipse Ti2) with a 100x oil immersion objective and a DS-QI2 camera (Nikon) at $15~\%$ relative DIA-illumination intensity and $100$~ms exposure time. The recording procedure has been performed for $800$ minutes, leading to a total of $4,000$ recorded microscopy images. The recorded images provide a spatial resolution of $0.072$ micrometers per image pixel in both spatial dimensions.

\subsection{Semi-automated segmentation \& tracking annotation}

To provide high quality annotation for the large number of recorded images with a limited amount of manual annotation workload, we decided to first perform an initial segmentation and tracking using Omnipose \cite{cutler_omnipose_2022} and UAT \cite{theorell_when_2019}, respectively. The result was subsequently corrected by an expert in the annotation tool ObiWan-Microbi~\cite{seiffarth_obiwan-microbi_2024}. For cell segmentation, we focused on providing annotations for every single-cell. Therefore, over- and under-segmentation, false positive and false negative segmentations were corrected. Based on the corrected segmentation, the tracking edges were manually checked and corrected. Table \ref{tab:man_corrections} shows the number of manual corrections carried out for the different time-lapse recordings. Only few manual segmentation corrections were performed. For cell tracking, the majority of corrections had to be performed towards the end of the time-lapse sequences, where the cell count and divisions events increased substantially and cells leave the field of view at the left and right image borders.

\begin{table}[]
    \centering

    \caption{Amount of manual tracking correction actions to correct errors in the semi-automated annotation workflow. A correction action is the addition and deletion of a tracking link or adding, deleting and editing of segmentation masks.}

    \begin{tabular}{@{} *{7}{c} @{}}
        \headercell{Manual\\Correction} & \multicolumn{6}{c@{}}{Time-lapses}\\
        \cmidrule(l){2-7}
        & 0 &  1 & 2 & 3 & 4 & Total  \\ 
        \midrule
          Segmentation  &  143 & 319 & 202 & 1087 & 25 & 1,776 \\ \midrule
          Tracking  & 773 & 1,463 & 4,057 & 1,823 & 1,111 &  9,227 \\ \midrule
        \end{tabular}

    \label{tab:man_corrections}
\end{table}

\subsection{Dataset statistics}

\begin{figure}
    \centering
    \includegraphics[width=\linewidth]{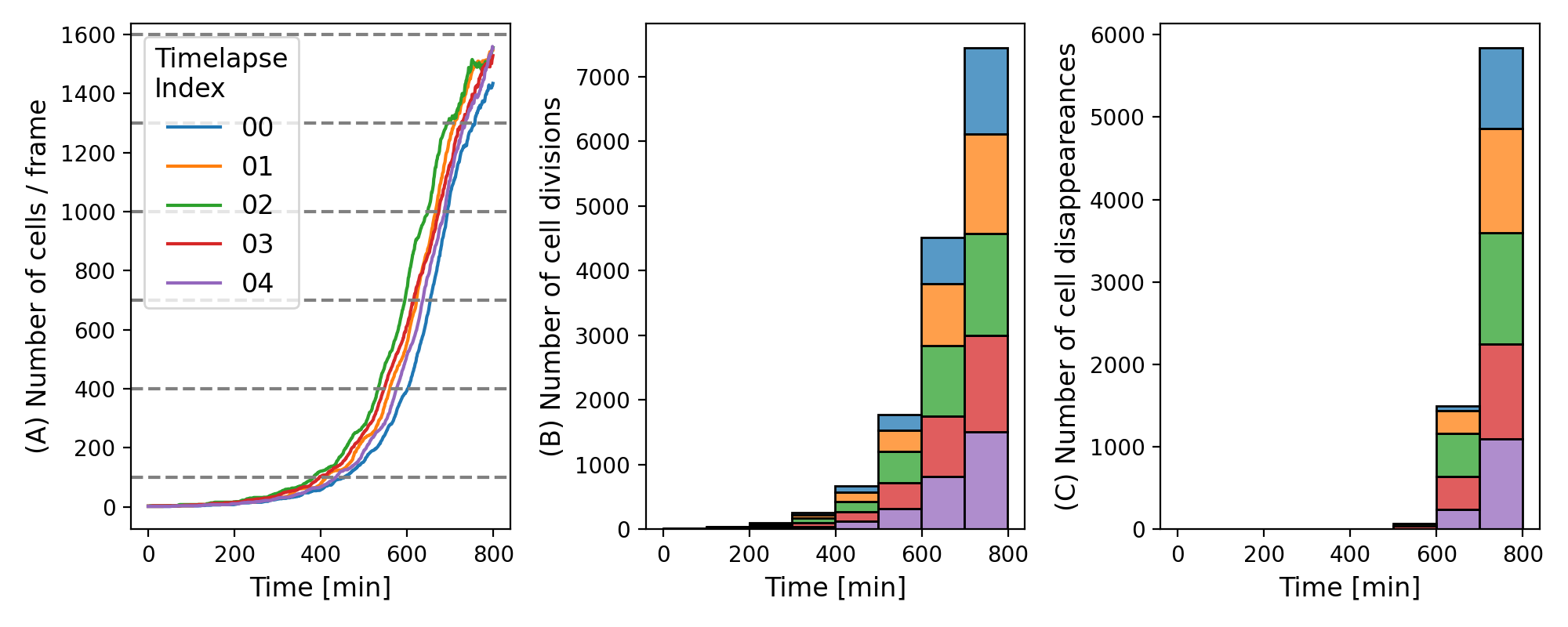}
    \caption{Temporal imbalance of the five microbial time-lapse recordings of the benchmark dataset. We measured the temporal development of cell count (A), cell divisions (B), and cell disappearance events (C) for each time-lapse. Cell division and disappearance events are grouped into bins of 100 minutes. The dashed lines in (A) indicate cell count limits (100, 400, 700, 1,000, 1,300, 1,600).}
    \label{fig:dataset_stats}
\end{figure}

The annotated dataset contains more than $1.4$ million densely annotated cell instances, $29$k cell tracks, and $14$k cell divisions (see Table~\ref{tab:data_set}). Thus, our data set is large enough to be split into meaningful train, validation, test splits.
Moreover, Figure \ref{fig:dataset_stats} shows that all five time-lapses show similar temporal developments of cell numbers, divisions and disappearances. 
Nevertheless, cell count and, thereby, the number of cell divisions are exponentially increasing throughout the time-lapses (Figure~\ref{fig:mlci}E) leading to a temporal imbalance. For instance, the large number of cells at the end of the time-lapses leads to $50~\%$ of the overall division events occurring in the last $100$ minutes of the time-lapse recordings. Moreover, the number of cell disappearances is much higher towards the end of the time-lapse recordings as the colony exceeds the size of the cultivation chamber and cells leave the field of view (Figure~\ref{fig:dataset_stats}B, C).

\begin{table}[]
    \centering
    \caption{Statistics for five time-lapse sequences in the benchmark dataset and its split into train, validation and test sets. The table shows the number of densely annotated segmentation masks (cell instances), cell tracks and cell divisions.}
    \vspace{0.2cm}
    \begin{tabular}{|c|c|r|r|r|r|}
        \hline
        Split & \makecell{Time-lapse\\Index} & \# Images & \# Cell Instances & \# Cell Tracks & \# Cell Divisions \\ \hline \hline
        Train & 0 & 800 & 238,364 & 4,918 & 2,448\\ \hline
        Train & 1 & 800 & 292,070 & 6,137 & 3,053\\ \hline
        Train & 2 & 800 & 327,832 & 6,884 & 3,428\\ \hline
        Val & 3 & 800 & 292,995 & 6,184 & 3,064\\ \hline 
        Test & 4 & 800 & 264,011 & 5,740 & 2,844\\ \hline \hline
        \textbf{Total} & \textbf{0,1,2,3,4} & \textbf{4,000} & \textbf{1,415,272} & \textbf{29,863} & \textbf{14,837} \\ \hline
        Train & 0,1,2 & 2400 & 858,266 & 17,939 & 8,929\\ \hline

    \end{tabular}
    \label{tab:data_set}
\end{table}

\subsection{Implications of imaging interval subsampling}

The temporal imbalance of microbial time-lapse datasets in Figure~\ref{fig:dataset_stats} is amplified by the choice of the imaging interval. 
Recording the TOIAM dataset with a low imaging interval allows us to simulate higher intervals by considering only every $k$th recorded image. We call this subsampling with a factor $k \in \mathbb{N}$ in the following. Due to the imaging interval of one minute, a subsampling of $k$ leads to an imaging interval of $k$ minutes. Using the subsampling procedure, we created subsampled datasets with higher imaging intervals.

\begin{figure}
    \centering
    \includegraphics[width=\linewidth]{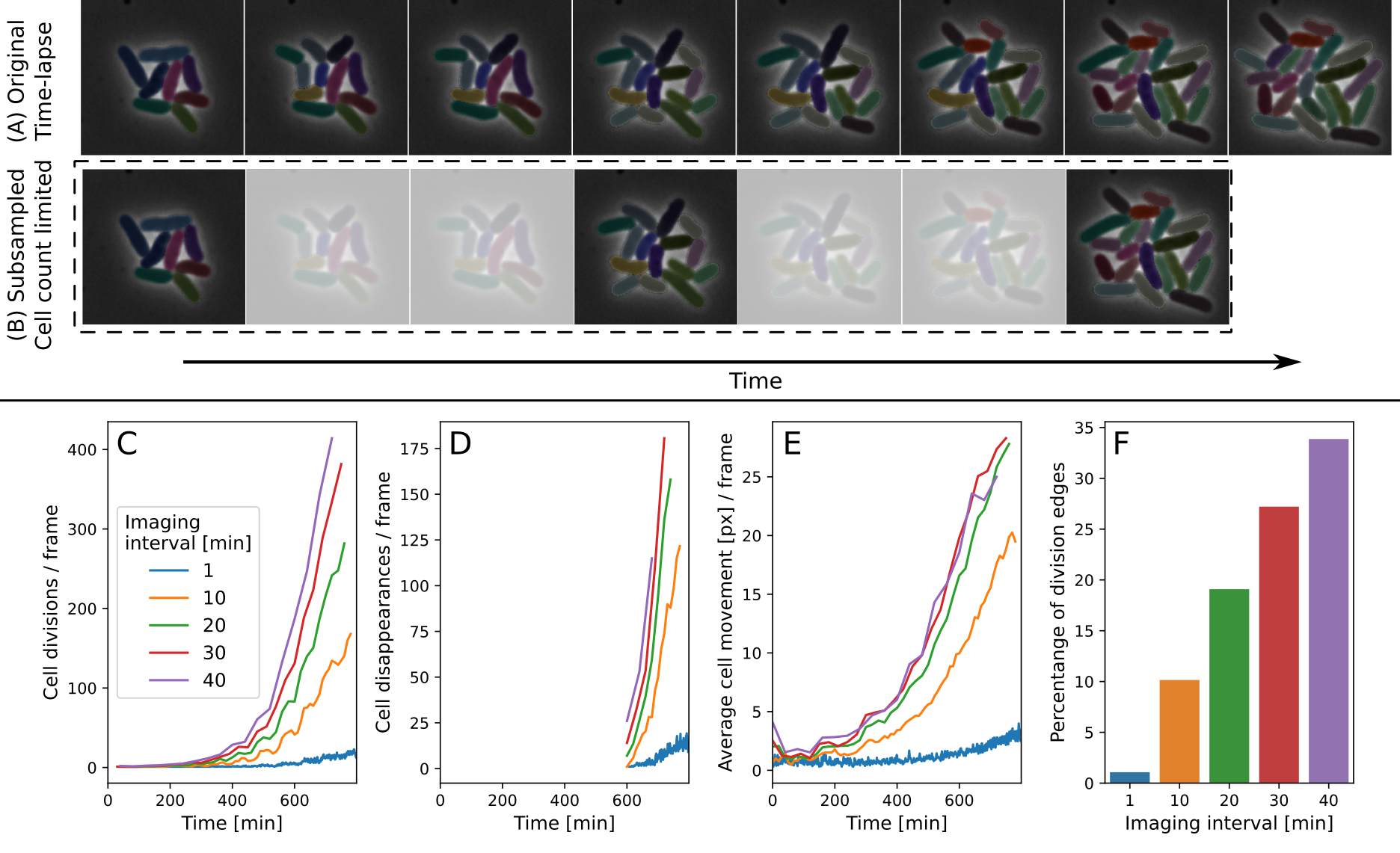}
    \caption{\textbf{Subsampling and cell count limiting of time-lapse sequences}. (A) shows an excerpt of an MLCI time-lapse. (B) shows an exemplary subsampling with a factor of $3$ and truncation at a cell count limit of $21$ leading to $3$ frames in total. Grayed out images denote frames removed due to subsampling, the dashed box denotes the cell count limit. (C-E) shows the temporal changes in the number of cell division, disappearance and movement per microscopy frame when subsampling to different imaging intervals. The curves have been exponentially smoothed. (F) shows the percentage of division events in contrast to non-division links. (C-F) show data from the TOIAM test split.}
    \label{fig:subsampling}
\end{figure}

Figure \ref{fig:subsampling} shows that changing the imaging interval has a tremendous impact on the structure of the dataset and, therefore, on the challenge to track the living cells. Time-lapses with simulated higher imaging intervals contain fewer images, but the number of cell divisions and cell appearances stays constant. Thus, the number of cell divisions and disappearances between consecutive frames increases strongly with higher imaging intervals (Figure~\ref{fig:subsampling}C, D). More cell divisions between consecutive frames lead to much larger cell displacements due to the 'snapping' cell division of \emph{C.~glutamicum} (see Figure~\ref{fig:subsampling}E). However, also the frequency of cell divisions increases. While at an imaging interval of one minute roughly $1~\%$ of cell links are cell divisions, higher imaging intervals lead to a strong increase, with up to $34~\%$ of the cell links being divisions at an imaging interval of $40$ minutes (\cref{fig:subsampling}F). 
Thus, having a good estimate on the frequency of cell divisions, for example based on previous experiments, is crucial.

\section{Experiment-aware microbial live-cell tracking metrics}

For a benchmark, suitable metrics are essential that rate the quality of the method performance and serve as an objective comparison tool. However, we have shown that MLCI time-lapses have unique characteristics that make the application of existing metrics difficult. Therefore, we present two new metrics that build on top of the well-established CTC metrics but introduce experiment awareness: First we incorporate the influences of experiment parameter choices, i.e. the imaging interval and maximum number of cells. We decompose the existing metrics along these experiment parameters and term these experiment-aware tracking metrics (\textit{EATM}). Second, we summarize the performance of tracking metrics across a wide range of these parameters within a single robustness metric (\textit{RM}).

\subsection{Cell tracking metrics}

The CTC introduces several metrics for measuring the quality of a tracking prediction in comparison with ground truth information \cite{ulman_objective_2017}. The \textit{TRA} and \textit{LNK} metrics are based on the Acyclic Oriented Graph Matching (\textit{AOGM}) score that determines the minimum number of operations needed to convert a predicted tracking result into its corresponding ground truth \cite{matula_cell_2015}. These operations include corrections of over- and under-segmentation and the addition or removal of tracking links. Each of these operations is awarded a constant penalty cost, and the AOGM gives their weighted sum. 

The \textit{TRA} metric is the \textit{AOGM}, normalized to $[0,1]$:
\begin{equation}
    TRA = 1 - \frac{\min\left( AOGM, AOGM_0 \right)}{AOGM_0} 
\end{equation}
where $AOGM_0$ is the $AOGM$ of an empty tracking graph (no nodes, no edges). 
The \textit{LNK} metric scores the quality of the tracking: 

\begin{equation}
    LNK = 1 - \frac{\min \left( \mbox{\textit{AOGMA}}, \mbox{\textit{AOGMA}}_0 \right)}{\mbox{\textit{AOGMA}}_0}
\end{equation}
where $\mbox{\textit{AOGMA}}$ denotes the $\mbox{\textit{AOGM}}$ where only edge operation costs are considered and the $\mbox{\textit{AOGMA}}_0$ denotes the $\mbox{\textit{AOGMA}}$ of the ground truth graph without edges, respectively.

Moreover, the \textit{DIV} and \textit{CT} metrics focus on correct reconstruction of 'biological events' that are cell divisions (\textit{DIV}) and complete cell tracks (\textit{CT}). Therefore, for both types of events true positives (TP), false positives (FP), and false negatives (FN) are computed, while precision and recall are summarized in the F1-score
\begin{equation}
    F_1 = \frac{2}{1 / precision + 1 / recall} = \frac{2 \cdot TP}{2 \cdot TP + FP + FN}.
\end{equation}

\subsection{\textit{EATM} cell tracking metric for MLCI}

Existing tracking metrics, such as those introduced by the CTC, have proven to be useful to rate the tracking quality. However, they do not consider the specific characteristics of microbial time-lapses that are crucially influenced by experiment parameter choices. Thus, we extend the existing metrics and decompose them in to the devised \textit{EATM} tracking metric, that considers the imaging interval and the maximal number of living cells.

Let $S=\{S_1, \ldots, S_L\}$ be the given segmentation ground truth 
of a time-lapse with $L \in \mathbb{N}$ frames. Then, some tracking-by-detection method $T$ predicts a tracking graph $G=(V,E)$ containing the segmentation detections as nodes, $V=\bigcup_{l \in \{1, \ldots, L\}} S_l$, and links between cell detections as tracking edges, $E \subseteq V \times V$.
We define a tracking metric to be a function $m(\cdot, \cdot)$ that compares a predicted tracking graph $\hat{G}=T(S)$ to the ground truth tracking graph $G^\star$ using a metric $m$ with normalized score:
\begin{equation}
    0 \leq m(\hat{G}, G^\star) \leq 1.
\end{equation}
The \textit{TRA}, \textit{LNK} and \textit{DIV} metrics defined before are such metrics.

First, we make such a metric sensitive to experiment parameters. Therefore, we evaluate the metric on temporally subsampled and cell count limited versions of the time-lapses (Figure~\ref{fig:sub_div_results}). 
Let $k \in \mathbb{N}$ be the subsampling parameter for reducing the temporal resolution. Let $N_{max} \in \mathbb{N}$ be the cell count limit. Then a time-lapse is truncated to the last frame where the cell count does not exceed the limit $N_{max}$~(Figure~\ref{fig:subsampling}B). We denote the subsampled and truncated segmentation information with $S|^k_{N_{max}}$ and the ground truth tracking graph with $G^\star|^k_{N_{max}}$, respectively. Then we define the experiment-aware tracking metric (EATM) version $\Tilde{m}$ of $m$
\begin{equation}
    \Tilde{m}^k_{N_{max}}(T, S, G^\star) := m \left( T(S|^k_{N_{max}}), G^\star|^k_{N_{max}} \right)
\end{equation}
that evaluates the metric $m$ on tracking prediction for the subsampled and truncated segmentation $T(S|^k_{N_{max}})$ with the subsampled and truncated ground truth tracking $G^\star|^k_{N_{max}}$.

Second, we define a new robustness metric (\textit{RM}) to summarize the robustness of the tracking algorithm across a wide range of imaging intervals and cell count limits. We define a set of subsamplings $SF \subset \mathbb{N}$ and maximum cell counts $MC \subset \mathbb{N}$. The robustness metric \textit{RM} of a metric $m$ measures the normalized frequency that the \textit{EATM} of $m$ surpasses a given threshold $\vartheta \in [0, 1]$:
\begin{equation}
    RM(m, \vartheta, SF, MC) := \frac{1}{|SF| \cdot |MC|} \sum_{k \in SF} \sum_{mc \in MC} \mathbbm{1}\left[\Tilde{m}^{k}_{mc} (T, S, G^\star) \geq \vartheta\right],
\end{equation}
where $\mathbbm{1}[\,\cdot\,]$ is the indicator function.

\section{Tracking evaluation}

We evaluated the performance of tracking methods on our new dataset using the \textit{EATM} and \textit{RM} metrics. For the comparison, we selected three representative tracking methods: The \texttt{Distance} method provides a baseline using pure distance information for linking costs and predicting links using a greedy scheme. The \texttt{LAP} method uses mask overlap for linking costs and optimizes the linking cost between consecutive frames in an LAP~\cite{jaqaman_robust_2008}. The \texttt{Trackastra}~\cite{gallusser_trackastra_2024} method is the best performing tracker according to the current CTC leaderboard. In our evaluation, \texttt{Trackastra} represents the data-driven methods for predicting linking costs.

\begin{figure}
    \centering
    \includegraphics[width=\linewidth]{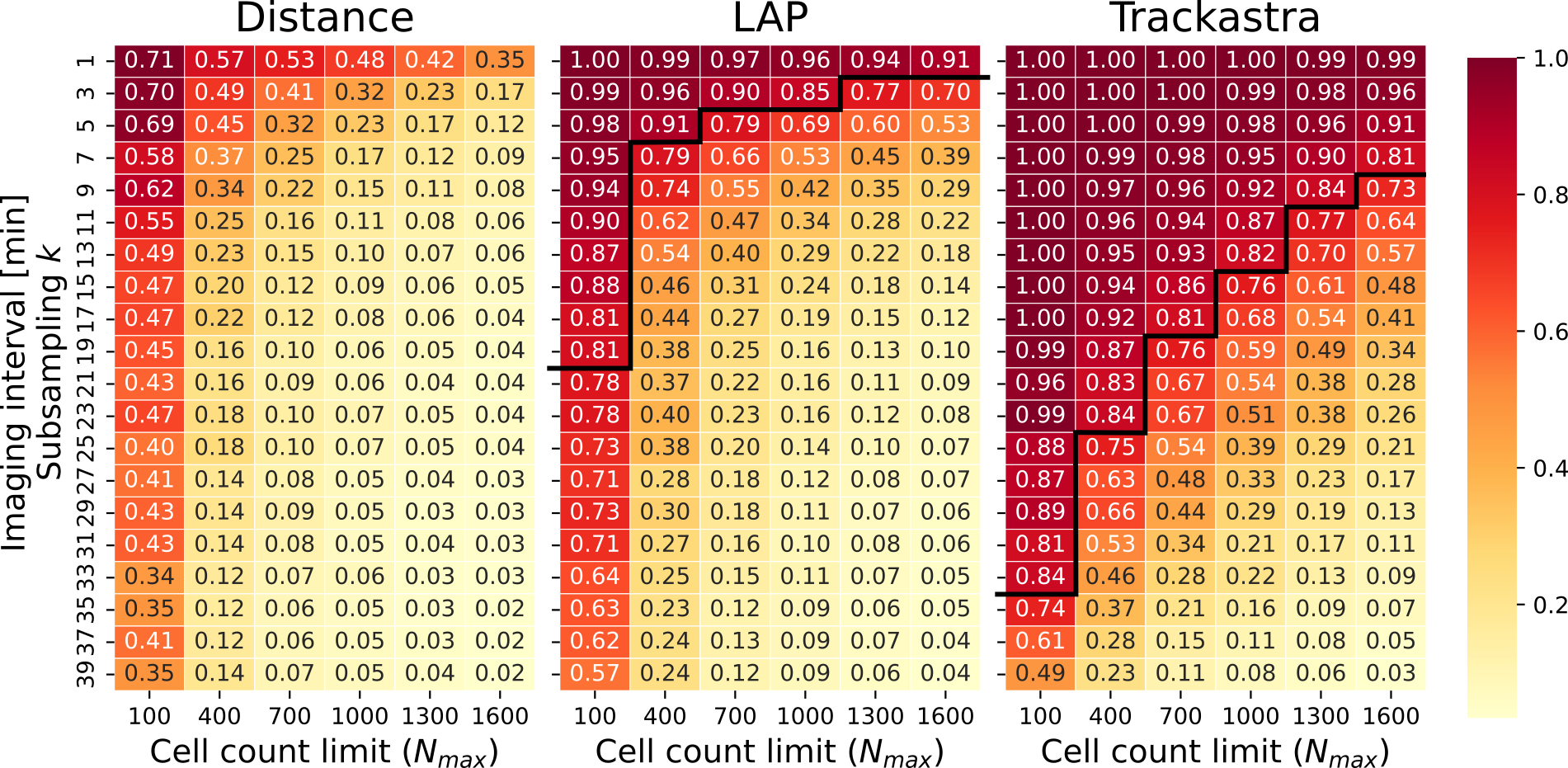}
	\caption{The \textit{EATM} based on the \textit{DIV} metric measured across different imaging intervals and cell count limits. The black line marks the region surpassing the $80~\%$ threshold ($RM@0.8$). The higher the value of the \textit{DIV} metric, the better is the reconstruction of the cell divisions, with a value of $1$ meaning perfect cell division reconstruction. Evaluations have been performed on the test split.}
    \label{fig:sub_div_results}
\end{figure}

We evaluated all three tracking methods with various imaging intervals and cell count limits on the TOIAM test split. Figure \ref{fig:sub_div_results} shows resulting heatmaps of the \textit{EATM} based on the \textit{DIV} metric.
Across all three tracking methods, the \textit{DIV} metric decreases 
for larger imaging intervals as well as higher cell limits, indicating deteriorating performance. 
The evaluation shows that higher cell numbers, more frequent cell divisions, and lower temporal resolution make the cell tracking task notably more difficult.

Among all three tracking methods, the \texttt{Distance} method performs worst across all experiment parameters. Therefore, making cell linking inferences based purely on distances, is not suitable for MLCI, especially when using a greedy scheme.
The \texttt{LAP} method performs slightly better in the $DIV$ metric at lower imaging intervals, benefiting from the more informative overlap costs and the non-greedy \texttt{LAP}. However, its performance drops rapidly when many cells are present and the imaging interval is increased leading to larger cell movement and, therefore, limiting the usefulness of overlap costs.
In contrast, \texttt{Trackastra} shows much stronger $DIV$ scores across the various experiment parameters. The transformer network trained on the training split seems to learn patterns that generalize to higher imaging intervals.
Thus, \texttt{Trackastra} robustly performs division reconstruction at various experiment parameters, opening the opportunity to monitor more cultivation chambers concurrently, which is important in the context of high-throughput MLCI-based screening applications.

\begin{table}[]
    \centering
	\caption{Robustness metric version of the \textit{TRA}, \textit{LNK}, and \textit{DIV} metrics evaluated on the test split using \textit{RM} thresholds of $80~\%$ and $90~\%$. The \textit{RM} metric has been computed over the subsampling factors ($SF$) and cell count limits ($MC$) used in Figure~\ref{fig:sub_div_results}.}
    \setlength{\tabcolsep}{6pt}
    \begin{tabular}{lrrrrrr}
        \headercell{Method} & \multicolumn{3}{c@{}}{RM@$0.8$} & \multicolumn{3}{c@{}}{RM@$0.9$}\\
        \cmidrule(l){2-4} \cmidrule(l){5-7}
        & \textit{TRA} $\uparrow$ & \textit{LNK} $\uparrow$ & \textit{DIV} $\uparrow$ & \textit{TRA} $\uparrow$ & \textit{LNK} $\uparrow$ & \textit{DIV} $\uparrow$ \\ 
        \midrule
        \texttt{Distance} (greedy) & \textbf{1.00}  & 0.16 & 0.00 & 0.46 & 0.12 & 0.00 \\ 
        \texttt{LAP} & \textbf{1.00} & 0.23 & 0.16 & 0.47 & 0.16 & 0.11 \\ 
        \texttt{Trackastra} (greedy) & \textbf{1.00} &	\textbf{0.50} & \textbf{0.45} & \textbf{0.81} & \textbf{0.42} & \textbf{0.32}
        \end{tabular}
    
    \label{tab:robustness}
\end{table}

While the \textit{EATM} heatmaps give detailed insights into the methods' performance, they only visualize a single tracking metric. To summarize and compare the methods across different metrics, we summarize the robustness of the tracking method to the experiment parameters using the \textit{RM} metric. Table~\ref{tab:robustness} shows the \textit{RM} score for the \textit{TRA}, \textit{LNK}, and \textit{DIV} metrics using a threshold of $80~\%$ and $90~\%$. For the \textit{DIV} metric, the \textit{RM} score is also visualized by the black line in Figure~\ref{fig:sub_div_results}.

We observe that the \textit{TRA} metric is not sensitive enough to give robustness insights when ground truth segmentation masks are provided. The correct segmentation will always lead to scores above $0.8$ in all experiment parameters, yielding a misleading robustness score of $1$.
The \textit{RM}s of \textit{LNK} and \textit{DIV} are more sensitive and provide insights for both robustness thresholds. \texttt{LAP} shows low results in \textit{LNK} and \textit{DIV} metrics, highlighting that it is only suitable for 
tracking cells at low imaging intervals and small cell colony sizes. The \texttt{Trackastra} method shows robustness up to $50~\%$ at an $80~\%$ threshold and, therefore, allows performing automated tracking at various experiment settings. Across all evaluations, the $DIV$ metric is consistently lower than the \textit{LNK} metric, indicating that cell division reconstruction is more difficult than linking non-dividing cells. This, underlines the larger focus on predicting cell divisions in MLCI datasets.

\section{Conclusions}

In this work, we have presented a new benchmark for cell tracking in MLCI with increased experiment awareness in metric ratings. The presented TOIAM dataset is the largest publicly available for MLCI in terms of annotated cell masks, cell tracks and cell divisions. Moreover, we have highlighted that MLCI data comes with unique challenges due to exponentially growing cell colonies and frequent cell divisions. These challenges are strongly influenced by experiment parameters such as the imaging interval and the maximum number of cells per frame. To capture these influences in appropriate metrics, we extended existing metrics towards experiment-awareness (\textit{EATM}) and summarize them in a robustness metric (\textit{RM}). We have shown that these \textit{EATM}s and \textit{RM}s give crucial insights into the practical suitability of tracking methods across a wide range of experiment parameters. Thus, our efforts aim to closely integrate method development and experiment design, and to open a stringent approach for 

experimenters to make informed decisions about their experiment parameter choices.

For now, our TOIAM dataset is limited to a single type of microbe cultivated within a single experiment. Therefore, we are looking forward to extending the dataset to other cell types and cultivation conditions and also apply the introduced metrics to other existing datasets in the future. Moreover, the evaluation of the tracking methods with erroneous segmentation and imperfect image data as well as the use of other biologically motivated metrics, such as the CT or BIO metrics from the CTC, is crucial for further evaluating their practical applicability. Adapting tracking methods to the challenging imaging conditions in MLCI, for example, by tuning hyper-parameters or establishing temporal subsampling during the training procedure might lead to more robust cell tracking methods.

Summarizing, our large-scale benchmark represents a step forward towards robust data-driven microbial single-cell tracking and facilitates tight integration of experiment parameters and tracking method development using experiment-aware metrics.

\section*{Acknowledgments}

This work was supported by the President’s Initiative and Networking Funds of the Helmholtz Association of German Research Centres [EMSIG ZT-I-PF-04-044]. JS and KN acknowledge the inspiring scientific environment provided by the Helmholtz School for Data Science in Life, Earth and Energy (HDS-LEE), and thank Wolfgang Wiechert for continuous support. NF, AJYS and RM were supported by the Helmholtz Program NACIP and the Helmholtz Information and Data Science School for Health (HIDDS4Health).

\bibliographystyle{splncs04}
\bibliography{eccv_2024}

\end{document}